\documentclass[sigconf]{acmart}

\setcopyright{none}
\renewcommand\footnotetextcopyrightpermission[1]{}

\AtBeginDocument{%
  \providecommand\BibTeX{{%
    \normalfont B\kern-0.5em{\scshape i\kern-0.25em b}\kern-0.8em\TeX}}}

\acmConference[]{}{}{}

\usepackage[mathcal]{eucal}
\usepackage{newtxmath}
\usepackage{algorithm}
\usepackage{algpseudocode}
\usepackage{subcaption}
\usepackage{indentfirst}
\begin{document}

\title{Incremental Object Grounding Using Scene Graphs}


\author{John Seon Keun Yi}
\authornote{Equal contribution}
\authornote{Georgia Institute of Technology, Atlanta, GA. Email: \{jyi62, ykim0606, chernova\}@gatech.edu}
\author{Yoonwoo Kim}
\authornotemark[1]
\authornotemark[2]
\author{Sonia Chernova}
\authornotemark[2]

\begin{abstract}
  Object grounding tasks aim to locate the target object in an image through verbal communications. Understanding human command is an important process needed for effective human-robot communication. However, this is challenging because human commands can be ambiguous and erroneous. This paper aims to disambiguate the human's referring expressions by allowing the agent to ask relevant questions based on semantic data obtained from scene graphs. We test if our agent can use relations between objects from a scene graph to ask semantically relevant questions that can disambiguate the original user command. In this paper, we present Incremental Grounding using Scene Graphs (IGSG), a disambiguation model that uses semantic data from an image scene graph and linguistic structures from a language scene graph to ground objects based on human  command. Compared to the baseline, IGSG shows promising results in complex real-world scenes where there are multiple identical target objects. IGSG can effectively disambiguate ambiguous or wrong referring expressions by asking disambiguating questions back to the user. 
\end{abstract}




\maketitle
\pagestyle{plain}

\section{Introduction}

In the context of human-robot interaction, object grounding is an important ability for understanding human instruction. In order for a robot to complete a task in response to a user's command, such as retrieving an object, the user must provide a description of the object with respect to the object's defining features and/or its relation to other surrounding objects. The object descriptions provided by the user are called referring expressions. 

Human referring expressions can be difficult for robots to understand because such phrases are often ambiguous or contain errors \cite{dougan2021open}.  Humans often utilize additional context or communication modalities, such as gestures or facial expressions, to resolve ambiguities \cite{holler2003pragmatic}.  However, today's robots typically lack the ability to process such inputs.

In this research, we aim to develop a system that can achieve object grounding in a human-robot interaction scenario. Our main contribution, Incremental Grounding using Scene Graphs (IGSG), is a disambiguation model that can clarify ambiguous or wrong expressions by asking questions to the human. We use scene graphs generated from both the referring expression and the image to obtain all relational information of the objects in the scene. In parallel, we use an off-the-shelf language scene graph parser \cite{schuster2015generating} to convert natural language to a scene graph since the linguistic structure of a language scene graph couples well with the image scene graph data structure and allows the grounding process to be incremental, pruning unnecessary computation. Grounding is achieved by incrementally matching the scene graph generated from the human command to the edges of the image scene graph. If needed, disambiguating questions are generated base on the relational information from the image scene graph.

Prior work has explored the use of natural language referring expressions and object detectors for object grounding \cite{hatori2018interactively}\cite{shridhar2018interactive}.  In this work, we contribute the first approach to leverage image scene graphs in this context, and demonstrate that the resulting semantic representation enables improved object disambiguation for complex scenes.  

We evaluate our model by testing on a set of images that contain multiple identical objects using clear, ambiguous, and erroneous referring expressions. Experiments show that our model outperforms prior state-of-the-art \cite{shridhar2018interactive} on interactive visual grounding on the tested images.
\par
Key contributions of our work include:
\begin{itemize}
    \item First work to utilize the linguistic structure of a language scene graph to process referring expressions.
    \item First work to use semantic relations from scene graphs for interactive visual grounding.
    \item First object grounding work to correct erroneous referring expressions from humans.
\end{itemize}

\section{Related Work}

In this section, we set the backbone of our work in scene graph generation by discussing relevant works on language scene graph parser and image scene graph generation. Then, we discuss previous works on object grounding and user command disambiguation to identify common problems and improvements implemented in our work. 

\subsection{Language Scene Graph}
A scene graph is a data structure that provides a formalized way of representing the content of an image. Due to its conciseness and its ability to represent a wide range of content, it was used by Johnson et al. \cite{johnson2015image} as a novel framework for image retrieval tasks where the authors used scene graphs as queries to retrieve semantically related images. The authors stated that a scene graph is superior to natural language because natural language requires a complex pipeline to handle problems such as co-reference and unstructured-to-structured tuples. Schuster et al. \cite{schuster2015generating} acknowledged the importance of the scene graph in image retrieval and presented a rule-based and classifier-based scene graph parser that converts scene descriptions into scene graphs. Our work utilizes the rule-based parser to generate language scene graphs from input referring expressions.

\subsection{Image Scene Graph}
Utilizing scene graphs in visual scenes enhance the understanding of an image much further than simple object detection. A visually grounded scene graph \cite{johnson2015image}\cite{xu2017scene} of an image can capture detailed semantics from an image by modeling the relationships between objects. The ability of the image scene graph to extract rich semantic information from images is being realized in visual question answering \cite{teney2017graph}\cite{hildebrandt2020scene}, image captioning \cite{yao2018exploring}\cite{yang2019auto}, visual grounding \cite{yang2018graph}, and image manipulation \cite{dhamo2020semantic}. While the promise of scene graphs is exciting, there has been difficulty extracting relational information efficiently and accurately. The naive approach of parsing through every possible relation triplets can be taxing with the increase of the number of objects in the scene. Yang et al. \cite{yang2018graph} attempts to resolve this problem by using a relation proposal network to prune irrelevant object-relation pairs and a graph convolutional network to capture contextual information between objects. Recently, a model proposed by Tang et al. \cite{tang2020unbiased} reduces the biased relations generated by traditional scene graph generation methods. \cite{zellers2018neural} Counterfactual causality of the biased model is used to identify bad biases, which are removed. This model is able to generate more detailed relationships between objects such as "standing on" or "holding" instead of simple relations like "on". We use the model from Tang et al. \cite{tang2020unbiased} to generate scene graphs from images.

\subsection{Object Grounding}
Object grounding is a task of locating an object referred by a natural language expression in an image. This referring expression includes the appearance of the referent object and relations to other objects in the image. Some approaches to tackle this task is to learn holistic representations of the objects in the image and the expression using neural networks \cite{zhuang2018parallel}\cite{yang2019relationship}\cite{yang2019cross}. However, these works neglect linguistic structures that could be utilized to disambiguate vague expressions. Other works use a fixed linguistic structure such as, subject-location-relation \cite{hu2017modeling} or subject-relation-object \cite{yu2018mattnet} triplets to ground an object which often fails in real world scenes which may need complex expressions. Yang et al. \cite{yang2020graph} uses neural modules to ground an object with the guidance of linguistic structure; namely the language scene graph extracted from a rule-based parser \cite{schuster2015generating}. Despite all the progress, the limitations of these grounding methods are that they assume perfect expressions and do not take into account vague or erroneous referring expressions. 

\subsection{Command Disambiguation in Robotics}  
While visual grounding models \cite{schuster2015generating}\cite{yang2020graph} assume that the given referring expression is clear and lead to eventual grounding, the same cannot be said with real-world human-robot interaction scenarios. Grounding objects in an environment where multiple objects match the user command requires further interaction to disambiguate the given task. For example, in Figure \ref{fig:eximage}, there are multiple cups in the image. A simple command to "grab the cup" will not lead to immediate grounding. The robot will need more detailed instructions, such as, "grab the green cup on the table" in order to achieve grounding.
\par
Several works attempt to resolve ambiguities by asking clarifying questions back to the human \cite{shridhar2018interactive}\cite{dougan2020impact}\cite{hatori2018interactively}\cite{magassouba2019understanding}. Li et al. \cite{li2016spatial} modeled abstract spatial concepts into a probabilistic model where explicit hierarchical symbols are introduced. Sibirtseva et al. \cite{sibirtseva2018comparison} compared mixed reality, augmented reality, and monitor as visualization modalities to disambiguate human instructions. Whitney et al. \cite{whitney2017reducing} adds pointing gestures to further disambiguate user command. Other works ask questions based on the perspective of the robot relative to the human \cite{li2016spatial}\cite{dougan2020impact} However, the results from these papers are generated in a carefully designed lab environment with limited variation of objects such as Legos \cite{sibirtseva2018comparison} or blocks \cite{sibirtseva2018comparison}\cite{paul2017grounding} and are restricted to tabletop \cite{shridhar2018interactive} or four-box \cite{hatori2018interactively}\cite{magassouba2019understanding} settings. Although those can be a straightforward baseline environment, it is far from a real-world environment with objects in various places.
\par
Shridhar et al. \cite{shridhar2018interactive} comes close to our goal by achieving interactive visual grounding using a real robot in a tabletop environment. However, their works are limited to rather simple referring expressions heavily based on positional relations such as "on" and "next to". Also, when there are multiple identical objects in in the scene, their model chooses to point at the objects one by one, instead of asking further questions.
\par
Furthermore, all of the works above fail to resolve wrong referring expressions. For example, going back to Figure \ref{fig:eximage}, "grab the green banana" will fail since there is no banana in the scene. While \cite{marge2019miscommunication} identifies and categorizes these as "impossible to execute" commands, there is no attempt in correcting it. Our proposed model is applicable in real-world scenes with diverse objects and require few interactions. Our model is also able to handle vague or wrong referring expressions. 

\begin{figure*}
  \centering
  \includegraphics[width=\textwidth]{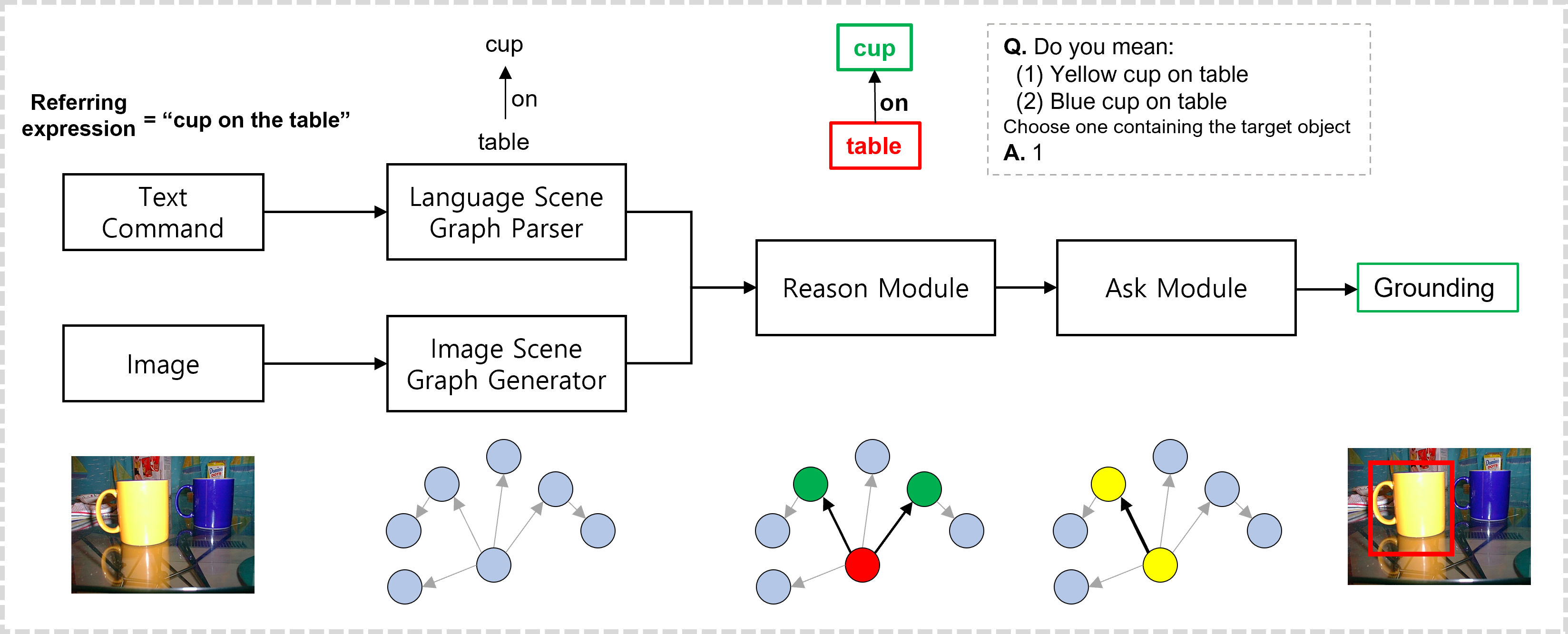}
  \caption{Overview of the Incremental Grounding using Scene Graphs. (IGSG) Different colors represent different nodes in the image and language scene graph, and bold arrows represent edges connecting two object nodes. Language and image scene graphs are constructed with the text command and input image. Then, the reasoning module matches edges from the image scene graph to one or more edges in the language scene graph. The generated list of edge candidates are then taken as input for ask module to ask disambiguating questions. Grounding is achieved when the human answers the questions.}
  \Description{model structure caption}
  \label{fig:model}
\end{figure*}

\section{Method}

Figure \ref{fig:model} represents the overall architecture of IGSG. Given an image and a referring expression of the target object in the image, IGSG grounds the target object and produces a bounding box around it. If needed, it can ask questions to clarify an ambiguous or wrong expression. This section provides a detailed explanation of the individual parts of IGSG.

\subsection{Overview}
IGSG achieves grounding using information obtained from the image and language scene graphs. Referring expression from a  user is taken as an input for the language scene graph parser and an RGB image is taken as an input for the image scene graph generator to produce scene graphs with nodes as objects and edges as relation between the objects. Figure \ref{fig:model} represents the overall architecture of IGSG. The two scene graphs are taken as input for the reasoning module. The reasoning module uses incremental matching to gather image scene graph edges that are similar to the language scene graph edges. The edge "candidates" generated from the Reasoning module is then carried over to the Ask module. The ask module asks questions based on the candidates. The human is asked to either select one of the options given or confirm a given relation. In case the candidate edges are identical, the ask module can ask distinctive questions for each candidate. Finally, object grounding is achieved based on the interaction between the human and the ask module.

\subsection{Language Scene Graph Generation}
Given a referring expression, we parse the command into a language scene graph using an off-the-shelf scene graph parser \cite{schuster2015generating}. The output language scene graph renders objects and their attributes as nodes and relations between the objects as edges. In this paper, we consider an edge from a scene graph as a relation "triplet": two object nodes and the relation connecting the two objects. A triplet consists of (subject, predicate, object), where the subject and object are nodes and the predicate is a relation. For example, "cup on table" would become (cup, on, table) in triplet form. 

Given a referring expression $\mathcal{L}$, we define the language scene graph representation of the expression as a set of edges $G^L=(E^L)$ where $E^L = (e^L_{1}, e^L_{2}, \ldots, e^L_{k})$. Each edge in $E^L$ consists of an subject node, the relation between the subject and object node, and the object node $e^L_{k}=(v^L_{k_s}, r^L_{k}, v^L_{k_o})$. All nodes in the set of $K$ nodes have the corresponding attribute of the object and its name: $V^L = \{A^L_{i}, N^L_{i}\}^{K}_{i=1}$. Therefore, a node in set of nodes will be $v^L_{k}=(a^L_{k},n^L_{k})$.

\subsection{Image Scene Graph Generation}
We use the image scene graph generator by Tang et.al \cite{tang2020unbiased} to extract scene graphs from images. This generator tries to avoid bias in predicate prediction stemming from skewed training data. Compared to other scene graph generators, this model can generate various unbiased predicates instead of common ones such as "on" and "has". 

Given an image $\mathcal{I}$, we define the image scene graph of the image $\mathcal{I}$ as $G^I=(E^I)$ where $E^I = (e^I_{1,2}, e^I_{1,3}, \ldots, e^I_{n-1,n})$ is the set of edges. Each edge $e^I_{i,j} = (v^I_i, r^I_ij, v^I_j)$ where $(i\neq j, i,j \in \mathbb{R})$ contains the subject $v^I_i$ predicate $r^I_ij$ and object $v^I_j$. The subject and object are from the set of $M$ nodes: $V^I = \{A^I_i, N^I_i\}^{M}_{i=1}$ where each node $v^I_i$ contains the object name $n^I_i$ and object attribute $a^I_i$.

\subsection{Reasoning Module}
The reasoning module takes each edge from the language scene graph and every edge from the image scene graph as input. It performs incremental matching to match the components from the language edge to the edges from the image. The goal of the reasoning module is to prune edges from the image scene graph and leave one or more "candidate" edges that are similar to the human command from the language scene graph.

Algorithm 1 explains the edge matching process. An edge from the language scene graph $e^L_i$ is compared with the set of image scene graph edges $E^I$. The overall incremental matching order is $object \rightarrow subject \rightarrow predicate \rightarrow attribute$. This means that if matching the edges by its object is not enough, the module moves on to matching the subject, then the predicate, and attribute. The $matchObject$, $matchSubject$, $matchPredicate$, and $matchAttribute$ functions in Algorithm 1 return edges from the image scene graph that contain the object/subject/predicate/attribute from $e^L_i$. For example, if $e^L_i =$ \textit{"cat on the table"}, $matchObject$ will return edges from $E^I$ that contain the object \textit{"table"}, such as \textit{"white plate on the table"}, or \textit{"lamp next to the table"}. 

Algorithm 2 shows the $matchObject$ function. It takes as input an edge from the language scene graph $e^L_i$ and multiple edges from the image scene graph. Notice that for  $e^L_i.object \approx e^I_k.object$, we use the Sentence-BERT \cite{reimers2019sentence} sentence embedding model to compare the cosine similarity of the object word pair. A pair of words that surpass the similarity score of 0.8 is considered matching.

There exist three cases in edge matching: one match, multiple matches, and no match. When there are one or more matches, the reasoning module moves on to match the next component in the matching sequence. When there are no matching edges, the module stops and the edge candidates from the previous match are used to ask questions. One exception is when there are no matching edges in the object match sequence. In this case we move on to match the $subject \rightarrow predicate \rightarrow attribute$. 

Through the reasoning process, irrelevant edges from the image scene graph are pruned and only candidate edges that match or are similar to the human command remain. When there are multiple edges in the candidate list, they are asked back to the human in order to disambiguate the initial command. The $ask$ function in Algorithm 1 represents this process. This is further explained in the next section.

\begin{algorithm}[t]
  \caption{Incremental Edge Matching}\label{euclid}
  \begin{algorithmic}
    \Procedure{INCREMENTAL\_MATCHING}{$e^L_i$, $E^I$}
      \State $obj = matchObj(e^L_i, E^I)$
      \If{$len(obj == 0$}
        \State $sub = matchSub(e^L_i, E^I)$
        \If{$len(sub) == 0$}
          \State No Grounding
        \Else
          \State $pred =  matchPred(matchAttr(e^L_i, sub))$
          \State $ask(pred)$
        \EndIf
      \Else
        \State $sub = matchSub(e^L_i, obj)$
        \If{$len(sub) == 0$}
          \State $ask(obj)$
        \Else
         \State $pred = matchPred(e^L_i, sub)$
         \If{$len(pred) > 1$}
           \State $attr = matchAttr(e^L_i, pred)$
           \State $ask(attr)$
         \Else
           \State $ask(pred)$
         \EndIf
        \EndIf
      \EndIf
    \EndProcedure
  \end{algorithmic}
\end{algorithm}

\begin{algorithm}[t]
  \caption{Object Matching}\label{euclid}
  \begin{algorithmic}
    \Procedure{matchObj}{$e^L_i$, $E^I$}
      \State $matching\_edges = []$
      \For{$e^I_k$ in $E^I$}
        \If{$e^L_i.obj \approx e^I_k.obj$}
          \State $matching\_edges.add(e^I_k)$
        \EndIf
      \EndFor
      \State \textbf{return} $matching\_edges$
    \EndProcedure
  \end{algorithmic}
\end{algorithm}

\subsection{Ask Module}
A critical step in disambiguation is validating the possible groundings through asking.
This means that the focus should expand beyond improving the agent's ability to understand humans better, to allowing humans to understand the agent better. However, prior solutions \cite{shridhar2018interactive}\cite{hatori2018interactively} that simply list the objects are not sufficient or can be more confusing for humans to reply accurately. For example, if the command is to "grab the white plate" in a scene where three white plates exist, there should be a way to differentiate between them so that the human can understand which white plate is where. We use relations with the surrounding objects from the scene graph to pick different relations such as "white plate near the black cat", "white plate next to the lamp", "white plate next to the silver fork". We do this by extracting the lowest occurring relation from each object node, which is illustrated in Algorithm 3. Given that we have multiple identical candidate edges $E^C$, we look at all image scene graph edges connected to each subject node in the candidate edge. Each candidate subject will have a list of edges from image scene graph. We count how many times an image scene graph edge has occurred for all candidates. Then, each candidate picks an edge with minimum number of overlap.

\begin{algorithm}[t]
  \caption{Find Relation}\label{euclid}
  \begin{algorithmic}
    \Procedure{findRelation}{$E^C$, $E^I$}
        \State $candidate = []$
        \For {$e^C_i$ in $E^C$}
            \State $relation = []$
            \For{$e^I_k$ in $E^I$}
                \If{$e^C_i.subject $ in $ e^I_k$}
                    \State $relation.add(e^I_k)$ 
                \EndIf
            \EndFor
            \State $candidate.add(relation\_edges)$
        \EndFor
        \State $occurrence = []$
        \For {$c$ in $candidate$}
            \State $count\_occurrence = \{\}$
            \For {$e^I$ in $c$}
                \State $count\_occurrence[e^I] = count(e^I)$
            \EndFor
            \State $occurrence.add(count\_occurrence)$
        \EndFor
        \State \textbf{return} $min(occurrence)$
    \EndProcedure
  \end{algorithmic}
\end{algorithm}

\section{Experiments}
We evaluate IGSG against INGRESS \cite{shridhar2018interactive}, the leading prior interactive human-robot visual grounding tool. INGRESS uses LSTMs \cite{hochreiter1997long} to learn the holistic representations of the objects in a scene and follows an iterative rule to ask disambiguating questions. 

We used a set of images from the Visual Genome dataset \cite{krishna2017visual} to perform our evaluation due to physical access restrictions resulting from COVID-19 closures at the time of the research. Also, we conducted a user survey to collect referring expressions for target objects in the images. Using the answers from the user survey, we generated a set of referring expressions to evaluate our model and INGRESS. We evaluated IGSG using two metrics: i) number of interactions, and ii) success rate.

\begin{figure}
  \centering
  \includegraphics[width=0.6\columnwidth]{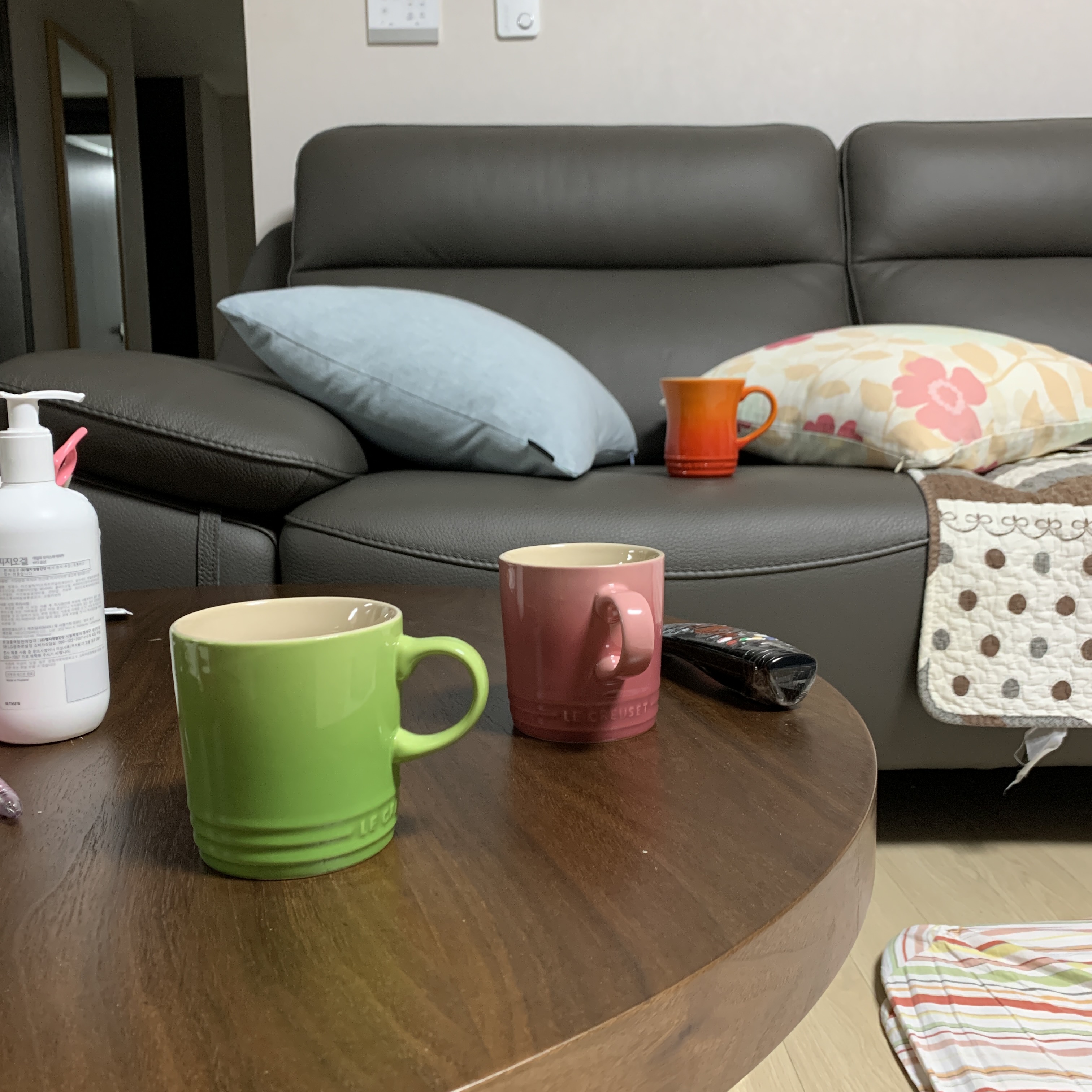}
  \caption{Example of an image used for object grounding. Three cups of different colors are in the scene, with two on the table and one on the sofa.}
  \Description{example image}
  \label{fig:eximage}
\end{figure}

\subsection{Settings and Data}
We selected fifteen images from the Visual Genome \cite{krishna2017visual} dataset to be used for object grounding. All selected images contain some components of ambiguity, meaning, multiple objects of the same class exist in the scene. These objects, such as cups or plates, may be of different color of shape, and are placed in the same or different places. For example, in Figure \ref{fig:eximage}, there are three cups in the scene. Simply asking \textit{"the cup"} or \textit{"the cup on the table"} is not enough to achieve grounding. There are an average of 9.66 objects per image, and the max number of identical objects in an image is 6. One of the objects in the scene is set as the "target object". The user has to give commands to the model to ground the target object. 

\subsection{Preliminary Experiment}
\begin{figure}[t]
  \centering
  \includegraphics[width=\columnwidth]{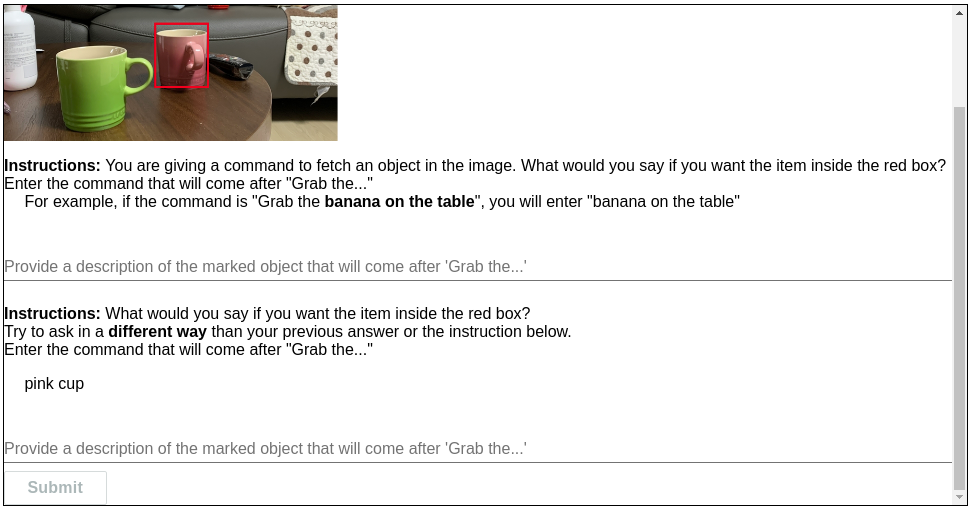}
  \caption{Question format used for the preliminary experiment. The participant is asked to write two different commands to ground the target object marked in the red box.}
  \Description{question format}
  \label{fig:questionformat}
\end{figure}

We conducted a user experiment in order to collect referring expressions from people without a robotics background in an actual human-robot interaction scenario. The goals of the experiment were to i) observe how people structure their commands in a scene with potential ambiguities, and ii) use the collected responses in the disambiguation experiment to avoid bias. The user experiment was done through Amazon Mechanical Turk. Figure \ref{fig:questionformat} demonstrates the question format seen by the test participant. Given an image, the target object is marked with a red box. The test participant is asked to provide two different commands in order for a robot to grab the target object. For the image in Figure \ref{fig:questionformat}, an example pair of commands would be "(grab the) cup next to the remote" and "cup next to the green cup". A total of 150 people participated in the experiment, and after eliminating irrelevant answers a total of 217 answers were collected.

The collected commands were divided into three categories: clear, vague, and not solvable by scene graphs. The first category contains commands that are clear meaning there exists only one object that fit the referring expression of the command. These commands does not require further disambiguation. The second category contains ambiguous or wrong commands that need to be clarified by asking questions. We consider a command is ambiguous if there exists more than one object that fit the referring expression of the command. The third category contains commands that are not solvable with semantic scene graphs. Commands in this category contain positional attributes like furthest, top, and third left. Since image scene graphs extract relations between a pair of objects, those that require comparison between three or more objects cannot be generated. Although such positional attributes does not exist in scene graphs, given a correct subject, our reasoning and asking module can disambiguate and ground the target.
Table 1 shows the distribution of categories of the collected commands. 

\subsection{Disambiguation Evaluation}
A total of 15 images were used for disambiguation evaluation.
Grounding commands were generated based on the responses collected from the preliminary experiment in section 4.2. Commands from all three categories (clear, vague, not solvable with semantic scene graphs) were used. In addition, complex commands that contain more than one edge were also used. An example of a complex command is \textit{"Black bag in the car next to the red bag"}. This command contains two edges \textit{"black bag in the car"} and \textit{"black bag next to the red bag"}. An average of 6.8 commands were used per image, and a total of 103 commands were used for evaluation.

For each evaluation, a ground truth target object is set and the grounding command is fed into the disambiguation model. When the model asks questions, a relevant answer is provided and the number of interactions is recorded. Grounding accuracy is assessed based on the final grounded object from the model. Grounding is successful when the grounded object is the target object.
\par
Both INGRESS and our IGSG use Faster-RCNN \cite{ren2015faster} for object detection. During the evaluation process, we noticed that the object detection pipeline fails to detect objects for some images. This behavior was especially noticeable on images with cluttered objects. Thus, we divide images into two batches based on the object detection results. Batch 1 contains images where Faster-RCNN successfully detects more than half of objects in the image. Batch 2 contains images where Faster-RCNN detects less than half of the objects in the image. Of the 15 images used, 10 images were in batch 1. We conduct disambiguation on batches and report results separately. 

\begin{table}
  \caption{Distribution of categories}
  \label{tab:freq}
  \begin{tabular}{ccl}
    \toprule
    Category&count\\
    \midrule
    clear & 83\\
    vague & 51\\
    not solvable & 83\\
    \bottomrule
\end{tabular}
\end{table}

\subsection{Results}
The experiment results are reported in Tables 2 and 3. As evaluation metrics we use i) Number of interactions, and ii) Success rate. The number of interactions is measured by the number of times the agent asks questions to the human. If the model achieves grounding immediately after the initial command, the number of interaction is zero. The success rate is assessed based on the final grounding result after interactions. IGSG is compared with INGRESS \cite{shridhar2018interactive} on the two metrics. We report results separately on the two batches of images.
\par
Table 2 illustrates experiment results on batch 1 images. For batch 1, the object detector gives correct bounding boxes and labels for most of the objects in the image. This allows the image scene graph generator to generate more solid semantic data for the model to utilize. The difference in access to this solid semantic data results in a significantly higher success rate compared to the baseline. 
\par
Table 3 illustrates the case where the object detector is not robust and fails to detect more than half of the objects in the image. With additional semantic data from the image scene graph generator, IGSG manages to improve the success rate compared to the baseline. Overall, the results show that IGSG outperforms the baseline in success rate with slight difference in average number of interactions regardless of the performance of the object detector. This shows that our model is more suited for real-world settings with complex scenes.
\par
In Table 4, to further analyze the success rate of our model, we divide the referring expressions into three categories used in the preliminary experiment section, and calculate the success rate of each category. This is done on batch 1 to disregard the effect of failing object detection as much as possible. IGSG significantly outperforms the baseline on all three categories, doubling the success rate when vague user commands were given. It is also notable that even though the image scene graph cannot fully process commands from the "not solvable" category, IGSG is able to achieve grounding in 80 percent of those referring expressions by interacting with the user. 
\par
\begin{figure}
  \centering
  \includegraphics[width=\columnwidth]{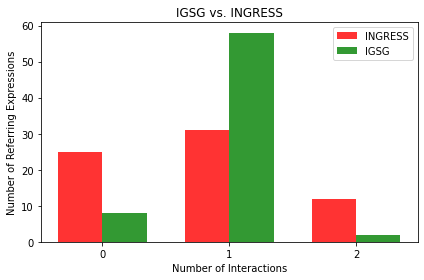}
  \caption{Histogram showing the number of interactions IGSG and the baseline used to ground an object for the provided referring expressions.}
  \label{fig:histogram}
\end{figure}

Figure \ref{fig:histogram} displays the distribution of the number of interactions for INGRESS and IGSG. Notice that IGSG has a high concentration of one interaction; the agent asks disambiguating questions to the human once. This is mainly because the asking module asks validating questions when there is a slight uncertainty in the reasoning module. For example, in the case of Figure \ref{fig:eximage}, if the provided referring expression is \textit{"green cup under the table"} and when the only candidate edge is \textit{"green cup on the table"}, the asking module still asks a question to the user to verify if the intended object is the green cup \textbf{on} the table. INGRESS tends to move straight to grounding without validation, which frequently lead to incorrect groundings. This behavior is reflected in the high number of zero interactions for INGRESS and its low success rate. Our model also rarely asks more than one question, and moves directly to grounding only when the referring expression exactly matches one edge from the image scene graph.

\begin{table}
  \caption{Grounding Results on Batch 1}
  \label{tab:freq}
  \begin{tabular}{ccl}
    \toprule
    Metrics&INGRESS&OURS\\
    \midrule
    Average Interactions & 0.809 & 0.89\\
    Success Rate & 0.441 & 0.779\\
  \bottomrule
\end{tabular}
\end{table}

\begin{table}
  \caption{Grounding Results on Batch 2}
  \label{tab:freq}
  \begin{tabular}{ccl}
    \toprule
    Metrics&INGRESS&OURS\\
    \midrule
    Average Interactions & 0.847 & 0.8\\
    Success Rate & 0.176 & 0.314\\
  \bottomrule
\end{tabular}
\end{table}

\begin{table}
  \caption{Success Rate by Category on Batch 1}
  \label{tab:freq}
  \begin{tabular}{ccl}
    \toprule
    Category&INGRESS&OURS\\
    \midrule
    clear & 0.432 & 0.757\\
    vague & 0.423 & 0.885\\
    not solvable & 0.6 & 0.8\\
  \bottomrule
\end{tabular}
\end{table}

\subsection{Examples}
\begin{figure}
    \centering
    \begin{subfigure}{0.5\textwidth}
      \centering
      \includegraphics[width=0.8\linewidth]{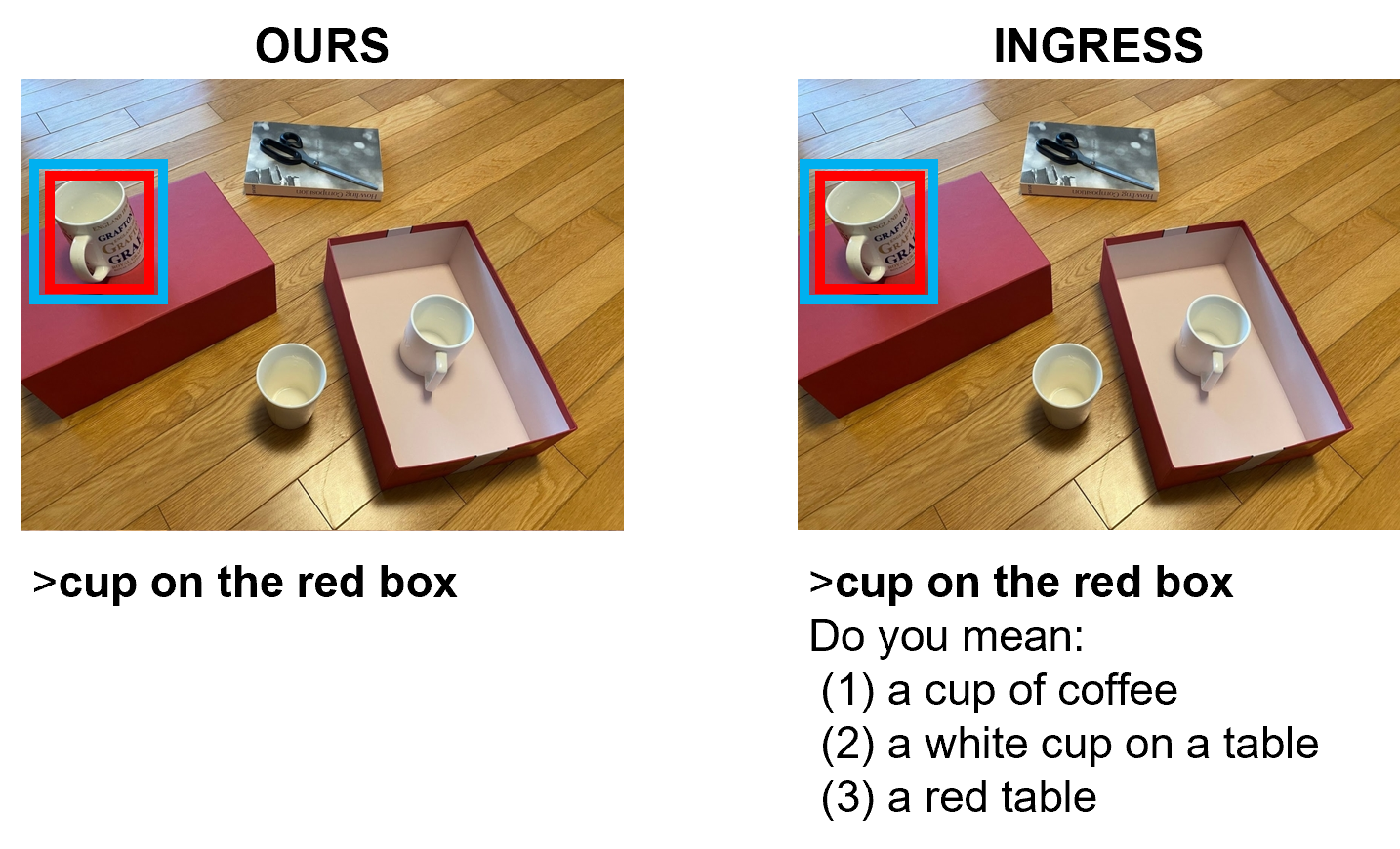}
      \caption{}
      \label{fig:sfig1}
    \end{subfigure}
    \begin{subfigure}{0.5\textwidth}
      \centering
      \includegraphics[width=0.8\linewidth]{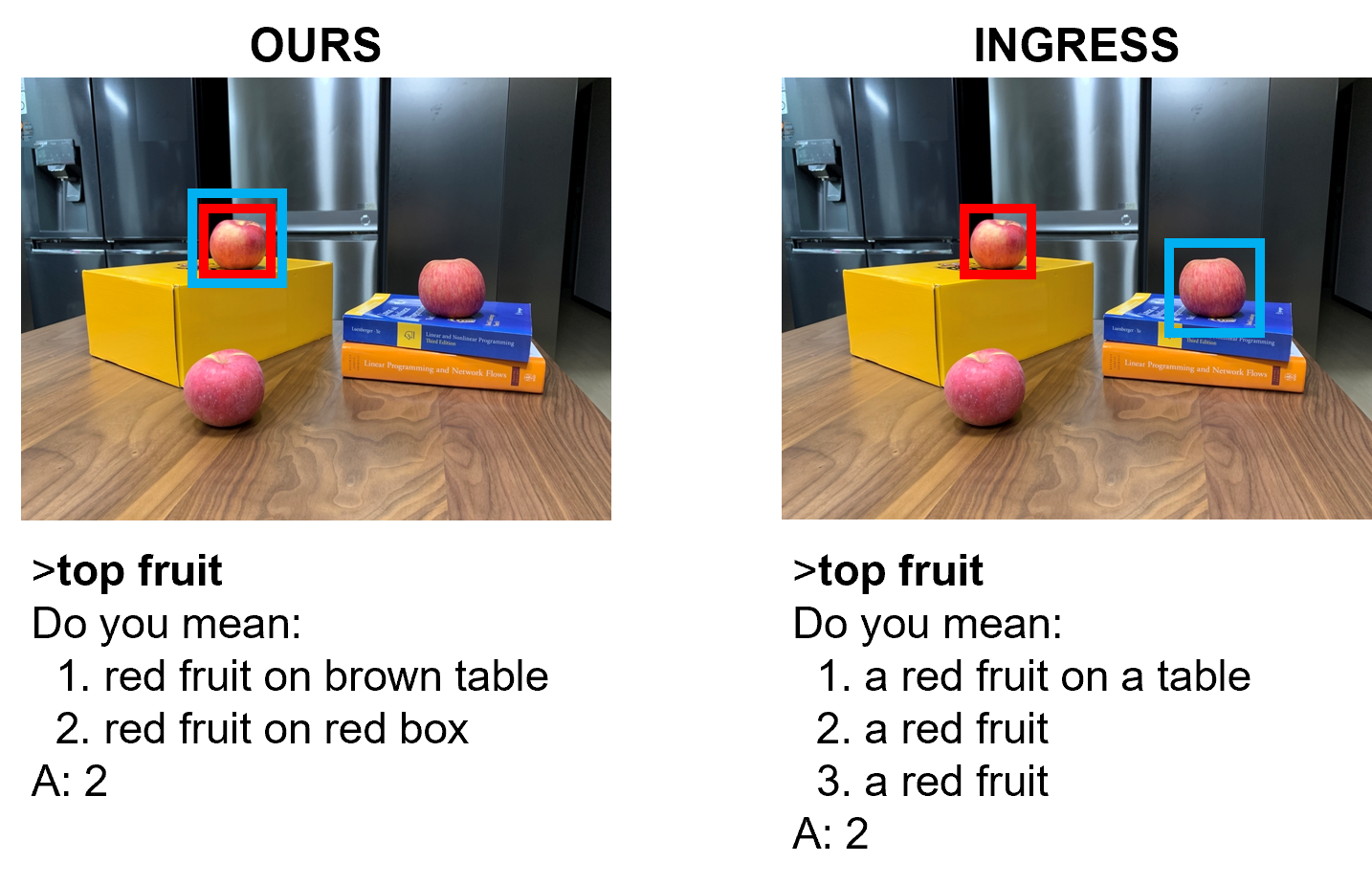}
      \caption{}
      \label{fig:sfig2}
    \end{subfigure}
    \begin{subfigure}{0.5\textwidth}
      \centering
      \includegraphics[width=0.8\linewidth]{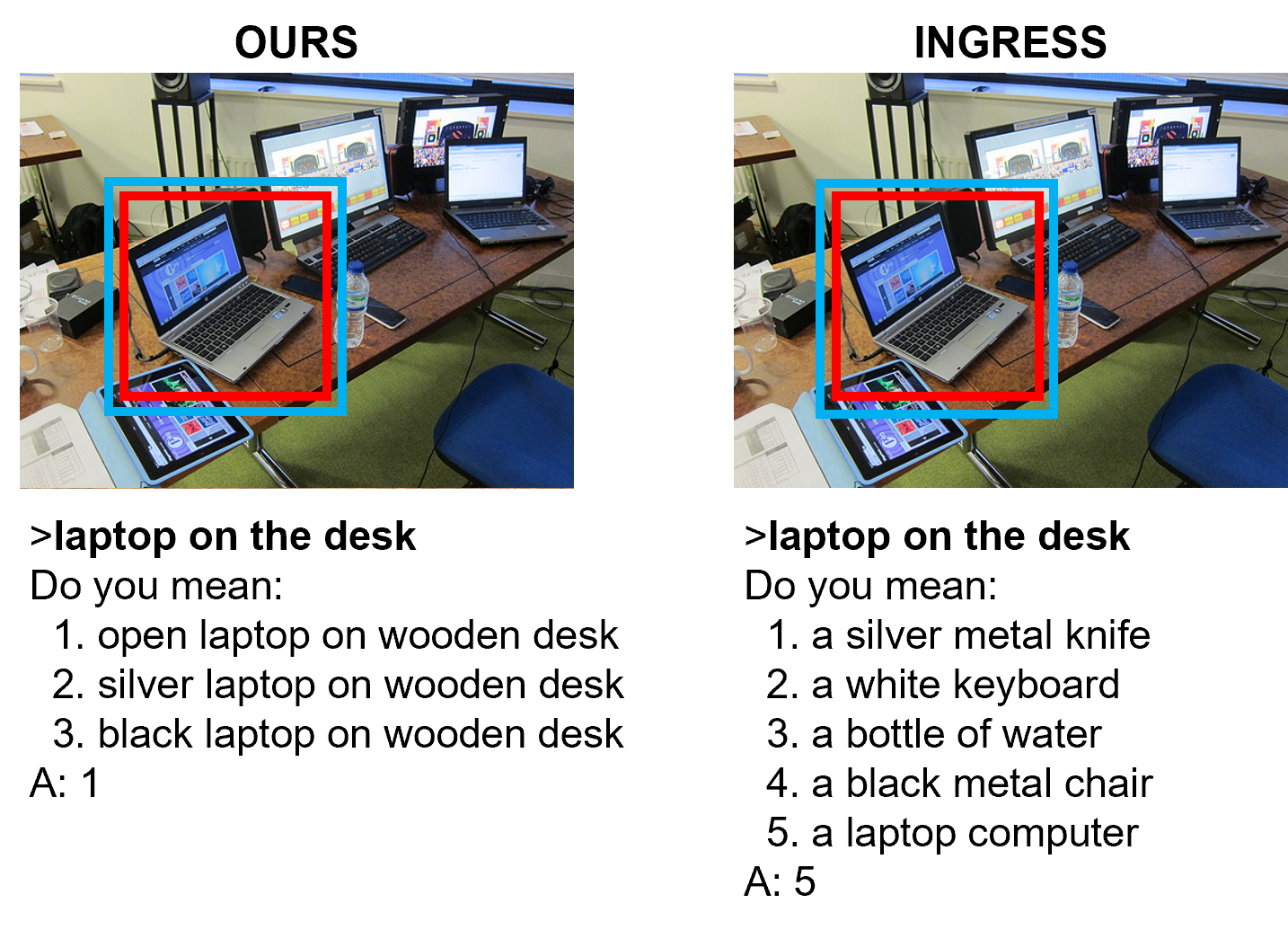}
      \caption{}
      \label{fig:sfig3}
    \end{subfigure}
    \begin{subfigure}{0.5\textwidth}
      \centering
      \includegraphics[width=0.8\linewidth]{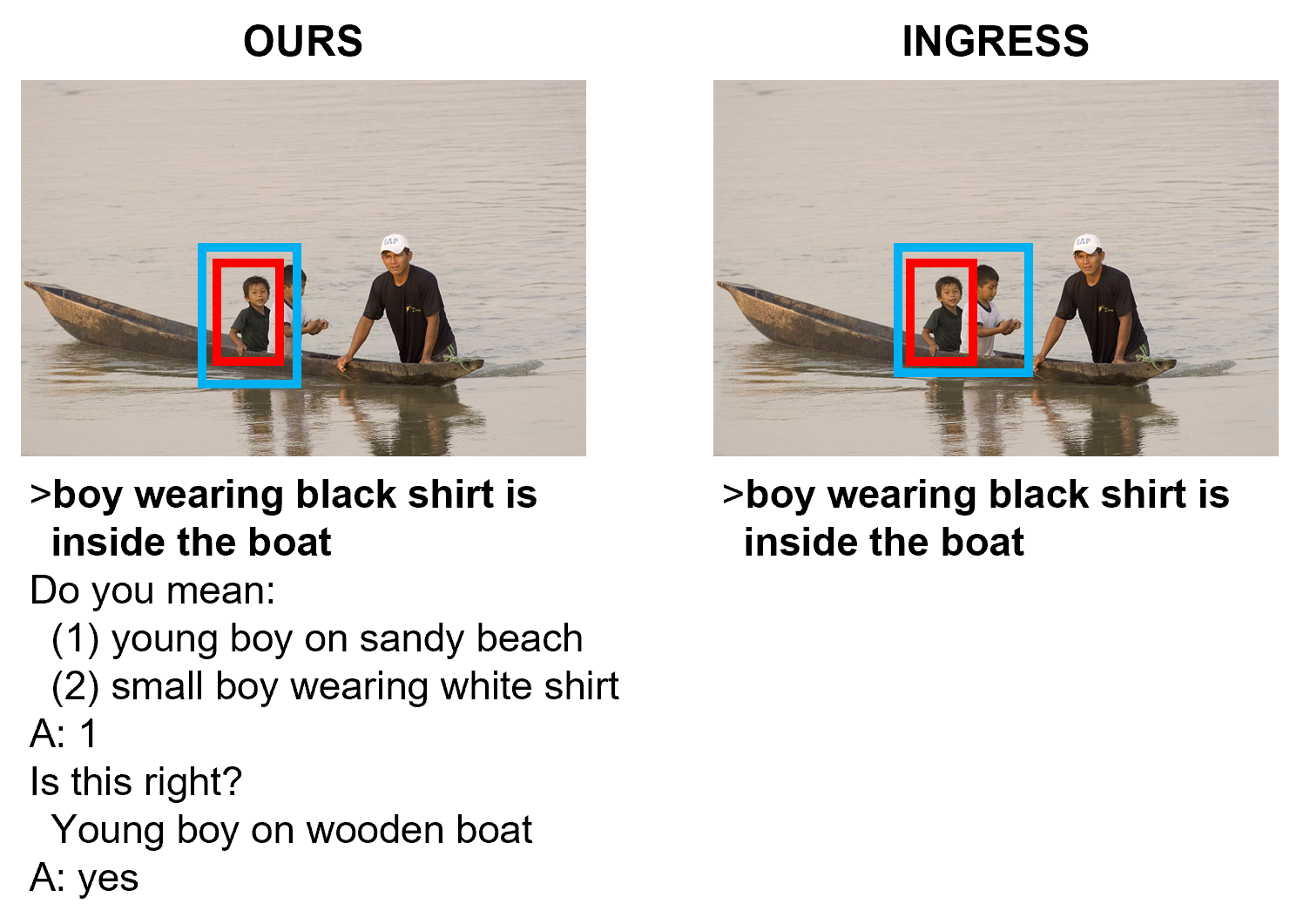}
      \caption{}
      \label{fig:sfig4}
    \end{subfigure}
    \caption{Examples of grounding results from our model (IGSG) and INGRESS \cite{shridhar2018interactive}. Objects inside the red boxes are the ground truth targets. Objects inside the blue boxes are grounded objects by the models. The text under the images show the initial human command and interactions.}
    \label{fig:results}
\end{figure}

Figure \ref{fig:results} shows some examples of the grounding results. The object inside the red box indicates the ground truth target object, and the object inside the blue box is the final grounded object. Figure \ref{fig:sfig1} represents a case where IGSG achieves direct grounding. Since there is only one cup on the box and the language referring expression matches one edge from the image scene graph, we can achieve grounding right away without asking further questions. However, INGRESS fails to identify the only cup on the box and has to go through one iteration of interaction to achieve grounding.
The referring expression used in Figure \ref{fig:sfig2} cannot be solved with a semantic scene graph, since the image scene graph cannot identify the positional information "top". However, IGSG can still achieve grounding by using the subject "fruit" and asking questions based on it. Notice the two questions asked (the object detection layer fails to detect the third fruit on the far right) contain distinctive relations of the two fruits. Questions asked by INGRESS, on the other hand, are not very useful since the human cannot distinguish the two "red fruit".
In the case of Figure \ref{fig:sfig3}, INGRESS asks questions that are irrelevant to the subject (laptop) from the referring expression. Only one of the questions is about the laptop. It also does not give any information about the relation of the objects.

The referring expression in Figure \ref{fig:sfig4} contains two edges: "boy wearing black shirt" and "boy inside the boat". IGSG asks questions for each edge. Since the "young boy" selected from the first and second iterations point to the same object, it achieves grounding after two interactions. INGRESS achieves direct grounding but fails to ground to the right object. 

\begin{figure}
  \centering
  \includegraphics[width=\columnwidth]{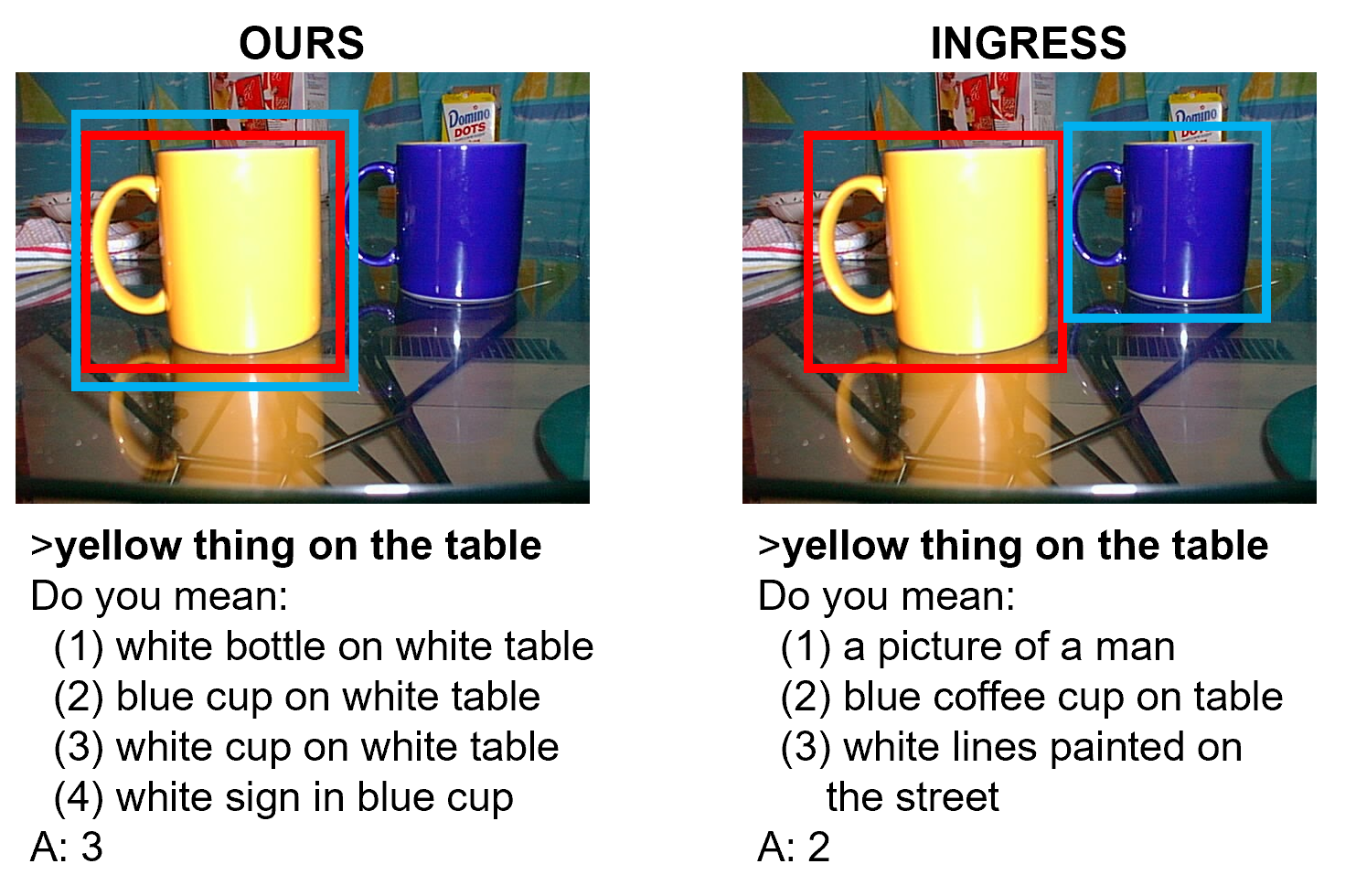}
  \caption{Example of a wrong referring expression. IGSG is able to achieve grounding even without the cup being mentioned. Here object detector incorrectly classifies the yellow cup as a white cup.}
  \label{fig:resultswrong}
\end{figure}

Figure \ref{fig:resultswrong} contains an experiment with a wrong referring expression. The input expression "yellow thing on the table" does contain information about the subject attribute (yellow) and its position, (on the table) but the subject "thing" does not exist in the image. While INGRESS fails to ask the edge containing the yellow cup, IGSG does and achieves grounding. 

\section{Discussion}
Although the experiment results show promising results, IGSG has some limitations. Mainly, false or failed predictions from the Faster-RCNN object detector and the scene graph generator results in a corrupted image scene graph. This prevents IGSG from achieving accurate grounding and asking relevant disambiguating questions. With an "ideal" image scene graph generated for the images used for evaluation, IGSG can achieve 0.68 average number of interactions and a success rate of 91.17 percent. Second, the robustness of the language scene graph parser is limited due to its rule-based approach. Since \cite{schuster2015generating} was first proposed, there has been recent advances in natural language processing with the appearance of LSTMs \cite{hochreiter1997long} and transformers \cite{vaswani2017attention}. We believe a data-driven learning based approach that encompasses the recent advancements of natural language processing, can further increase the robustness and accuracy of the language scene graph parser. 

\section{Conclusion}
In this work we presented IGSG, an incremental object grounding model using scene graphs. IGSG achieves grounding by incrementally matching the language scene graph generated from the human's referring expression to the image scene graph created from the input image. When the referring expression is ambiguous or wrong, IGSG can ask disambiguating questions to interactively reach grounding. Our model outperformed INGRESS \cite{shridhar2018interactive} in a visual object grounding experiment both in the number of interactions and success rate. Through this model, we presented a new perspective in object grounding where we acknowledge that humans can be ambiguous and can make mistakes when giving commands to robots.
Although existing limitations of scene graph generation models prevent us from creating a perfect grounding model, we hope this is a right step in the direction of using semantic information from scene graphs for grounding and disambiguation. 

\bibliographystyle{ACM-Reference-Format}
\bibliography{references}

\end{document}